\documentclass{article}

\usepackage{PRIMEarxiv}

\usepackage[utf8]{inputenc} 
\usepackage[T1]{fontenc}    
\usepackage{hyperref}       
\usepackage{url}            
\usepackage{booktabs}       
\usepackage{amsfonts}       
\usepackage{nicefrac}       
\usepackage{microtype}      
\usepackage{lipsum}
\usepackage{fancyhdr}       
\usepackage{graphicx}       
\graphicspath{{media/}}     
\usepackage{filecontents}
\usepackage{amsmath,amssymb,amsfonts}
\usepackage{algorithmic}
\usepackage{textcomp}
\usepackage{xcolor}
\usepackage[most]{tcolorbox}
\usepackage{framed}
\definecolor{shadecolor}{RGB}{0,200,230}
\usepackage{graphicx}
\usepackage{adjustbox}
\title{Entity Embeddings : Perspectives Towards an Omni-Modality Era for Large Language Models
}

\author{
  Eren Unlu \\
  Datategy SAS \\
  Paris, France\\
  \texttt{eren.unlu@datategy.fr} \\
   \\
   \AND
   Unver Ciftci \\
   MATYZ Institute of Mathematics and Artificial Intelligence \\
   Tekirdag, Turkey \\
   \texttt{unver.ciftci@matyz.org} \\
}

\begin{document}
\maketitle

\begin{abstract}
Large Language Models (LLMs) are evolving to integrate multiple modalities, such as text, image, and audio into a unified linguistic space. We envision a future direction based on this framework where conceptual entities defined in sequences of text can also be imagined as modalities. Such a formulation has the potential to overcome the cognitive and computational limitations of current models. Several illustrative examples of such potential implicit modalities  are given. Along with vast promises of the hypothesized structure, expected challenges are discussed as well.
\end{abstract}


\section{Introduction}

Recent advancements in capabilities of Large Language Models (LLMs) have paved the way for an era in the field of natural language understanding and generation. A natural extension to the LLM powered natural language understanding would be to incorporate other modalities such as imagery and audio to harness the inherent cognitive strength of language, which acts as an intricate bridge connecting various semantic concepts \cite{wang2022image}\cite{tang2023learning}. This type of a formulation not only allows for beneficial possibilities like extracting information out of given non-textual modalities to perform tasks, but also add new dimensions in LLMs’ cognitive prowess through training. The significance of learning from multi-sensory stimuli and interaction to reach human level cognition and beyond is well deliberated \cite{jurafsky2020proceedings}\cite{ghazanfar2006multisensory}\cite{patterson2007you}.

Though it is hard to draw a hard line on the taxonomy, we can define the first image caption generators and image generation models with text based conditioning as early attempts to meld natural language understanding with a different modality \cite{balaji2019conditional}\cite{zhang2018stacking}\cite{zhang2023adding}\cite{huang2022fastdiff}. Despite their adeptness and popularity, these architectures have focused on a single particular task on processing distinct modalities which limit their flexibility to achieve greater cognitive complexity. Apart from the inability to fuse information from different modalities effectively, another practical limitation is that the modality heads such as imagery expect fixed input size. More importantly, the number of different data points which can be fed for a modality is static, usually one.

Recently, few LLM architectures with multi-modal data processing capability have been proposed with promising results \cite{wu2023next}\cite{chen2023x}\cite{driess2023palm}. Different from conditioned generative models with cross-attention and early multimodal LLMs, these proposals aim to project data of any modality on the joint latent language space. The central idea of these proposed architectures can be summarized as “tokenization” of data from different modalities and utilization of these “modality tokens” as regular language tokens allowing interleaving with proper projection and training mechanisms. 

If the current transformer centric architectural framework continues to exist and prosper, we believe that one particular direction that research will follow is this “any-modal” tokenized approach. Incorporating multiple diverse modalities into a unified linguistic space presents advantages that transcend contemporary expectations, which can eventually provide substantial breakthroughs. We refer to this direction as an “omni-modal” era in LLMs and build certain perspectives on how training data organization, model development and usage might evolve.

In this paper : 
\begin{itemize}
  \item We discuss architectures of two recent models which conform to aforementioned generic tokenized, multi-modal context. Their advantages and limitations are compared. Promise of these models as a foundation for further omni-modal architectures is evaluated.
 
  \item  The term “entity embedding” is introduced, where “entities” refer to any piece of information of any modality that can be represented by a finite amount of interleavable tokens on the latent linguistic fabric. The potential to use pure textual semantic structures as a form of compressed modality is also considered. 
  
  \item The possibility of recursive and interconnected utilization of the entity embeddings is discussed. Several hypothetical use cases are given as illustrative examples. 
  
  \item We share our perspective and prospects of such an omni-modal era : possibilities, technical challenges and requirements.
\end{itemize}

\section{Tokenized Generic Multi-modal LLMs}

As mentioned previously, we refer to models which project representations of any modality onto the common latent space with linguistic tokens as generic “tokenized” architectures. Though, they may have not demonstrated the capability to use more than one modality currently, we consider them as generic architectures as they can potentially permit. As all modalities are represented as tokens in the aligned embedding space, they allow multi-modal in-context learning as well as regular LLMs. Though there exists various other studies, two of the most recent architectures are discussed in this paper :  AnyMAL and Kosmos-I. 

\textbf{AnyMal} : Any-Modality Augmented Language Model (AnyMAL) \cite{moon2023anymal} proposes to use projection layers for each modality as an extension to \cite{tsimpoukelli2021multimodal} and additional fine-tuning for multi-modal context. \cite{tsimpoukelli2021multimodal} approaches multi-modal task completion as a rather limited few-shot learning problem. The authors demonstrate capabilities only for visual data, however potentially the method is applicable to any modality. The central idea of \cite{tsimpoukelli2021multimodal} is to train modality specific encoders end-to-end while keeping the weights of the backbone language model frozen. This is somehow related to previously proposed prefix tuning and prompt tuning and a certain extension of these concepts in different modalities \cite{li2021prefix}\cite{lester2021power}. The modality data is encoded and tokenized with a specific encoder, which is an image model in this case. In order to align these modality specific tokens with the foundational LLMs’ embedding space, only the encoder weights are updated as illustrated in Fig. 1. This in turn shows promising emerging abilities in multi-modal few-shot learning. The authors have trained the vision encoder in this setting with only single image-text prompt pairs, however the model is still capable of few-shot learning even with a given series of multiple visual data points. 

\begin{figure}[h]
\centering
\label{fig:fig_1}
\includegraphics[width=0.65\linewidth]{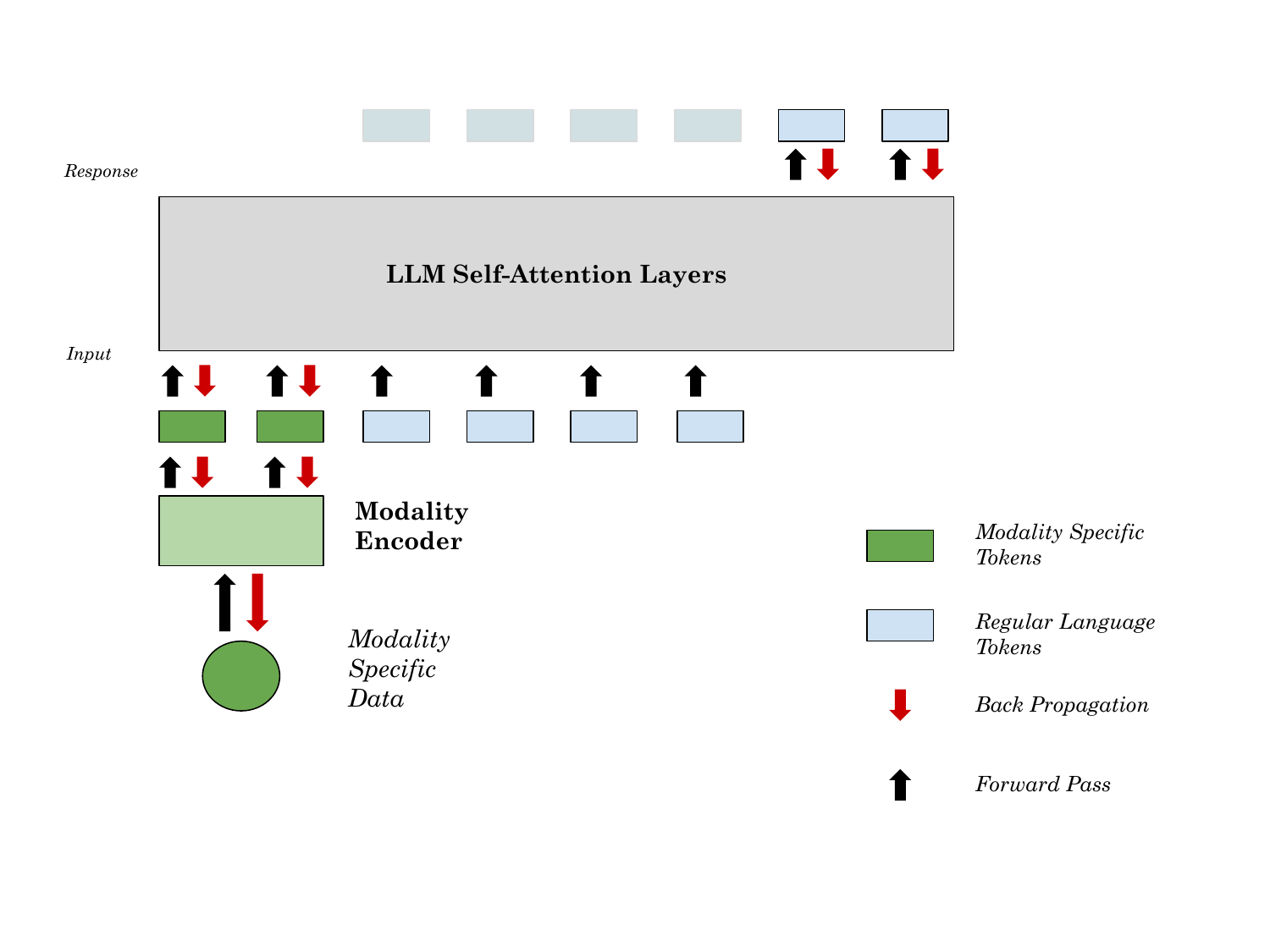}
\caption{\cite{tsimpoukelli2021multimodal} employs an encoder to produce modality specific tokens which is end-to-end trained while keeping all LLM layers frozen.}
\end{figure}

AnyMAL proposed by \cite{moon2023anymal} aims to bring such a setting into a more generic and cognitively capable level. For this purpose, projection layers are introduced in addition to modality encoders. First of all, modality encoders are chosen from available relatively large and capable pre-trained models with textual interaction such as CLIP for imagery and CLAP for audio \cite{schuhmann2022laion}\cite{robinson2023transferable}. Rather than retraining these encoders which are already aligned with textual data up to a degree, the authors employ learnable projection layers as in Fig. 2, which cast the outputs of encoders on the joint linguistic embedding space of the LLM. Different from \cite{tsimpoukelli2021multimodal}, the development of the model requires two stages of training : a pre-training phase to teach projection layers of each modality with an extensive dataset of modality-textual label pairs followed by an eloborate instruction tuning. The capabilities of the model are demonstrated for image, video, audio and even motion with Inertial Measurement Unit (IMU). A minor drawback of the architecture might be the static locations of instruction and modality tokens which limits the flexibility to manipulate multi-modal data linguistically.

\begin{figure}[h]
\centering
\label{fig:fig_2}
\includegraphics[width=0.65\linewidth]{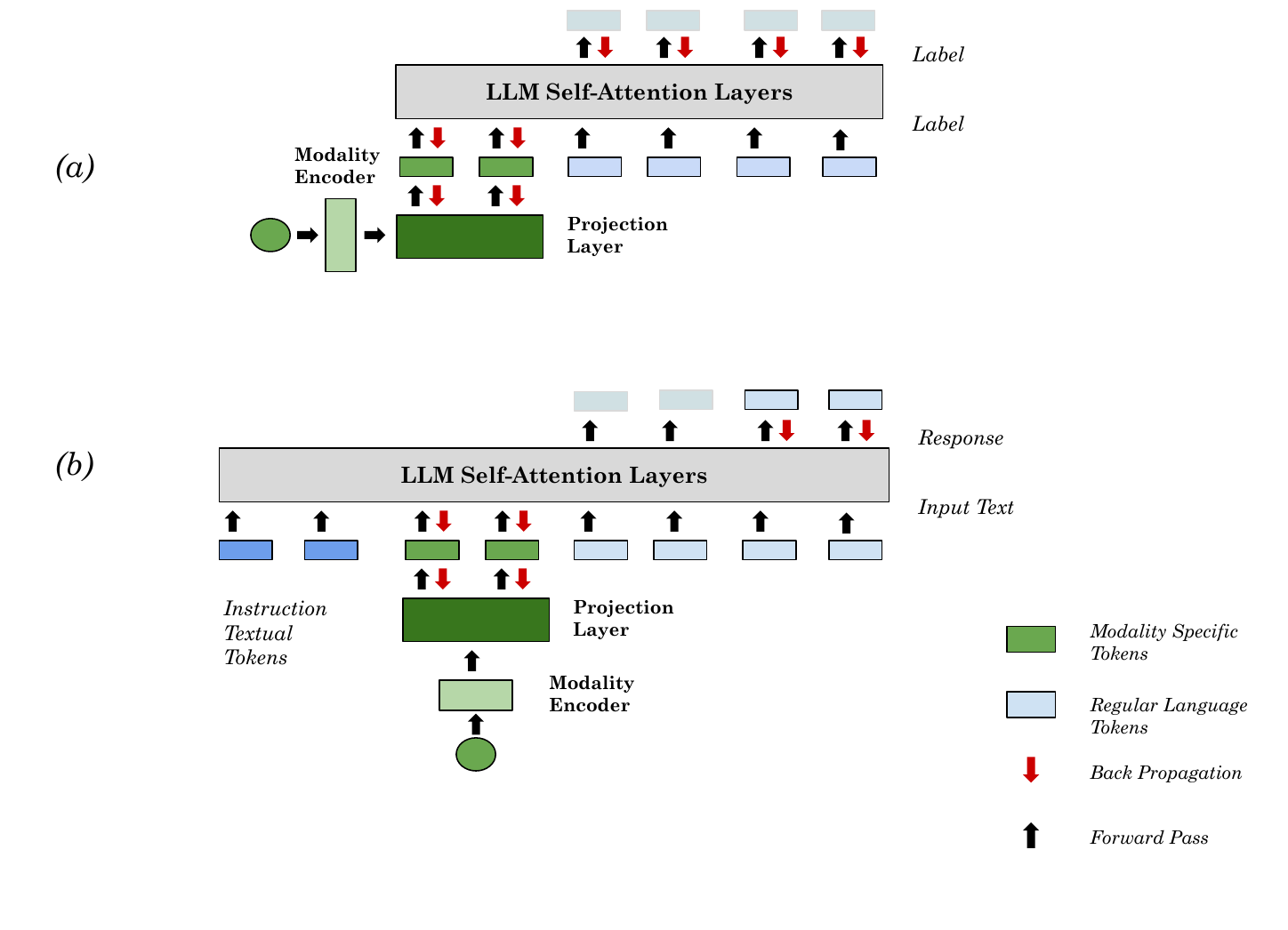}
\caption{(a) Proposed projection layers in \cite{moon2023anymal} where they are first pre-trained with image-caption pairs, while keeping the LLM and modality encoder layers frozen. (b) Instruction fine-tuning of the model.}
\end{figure}

\textbf{Kosmos-I} : \cite{huang2023language} approaches multi-modal intelligence from a more generic perspective. As in \cite{hao2022language}, every modality is represented with interleavable sequence of tokens as in Fig. 3. Like previously mentioned models, Kosmos-I also uses encoders to generate a set of tokens consumable with regular language tokens. Each sequence of modality tokens are decorated with special tokens at the beginning and the end. As it can be concluded, this straightforward approach is the most generic and modular architecture for omni-modality. However, It can be argued that the price to pay for such extensibility is to train the foundational language model from scratch, as trying to align multimodal embeddings in a flexible interleavable scheme with decorative tokens might not yield an adequate performance. The authors build a 24-layer generic transformer architecture with 32 attention heads allowing a context length of 2048 tokens. As mentioned previously, the model is trained from scratch with a diverse and vast multi-modal corpora. In their demonstration, the extra modality is only imagery. As such a generic approach allows to place textual and non-textual tokens without a specific pre-order, the training corpus mixed with images and text can be fed into the model without extra preprocessing. However, in order to further facilitate the training procedure, proper image-caption pairs are included in the corpus as well. Following extensive pre-training, the model is further fine-tuned and instruction aligned for multi-modal context.

\begin{figure}[h]
\centering
\label{fig:fig_3}
\includegraphics[width=0.65\linewidth]{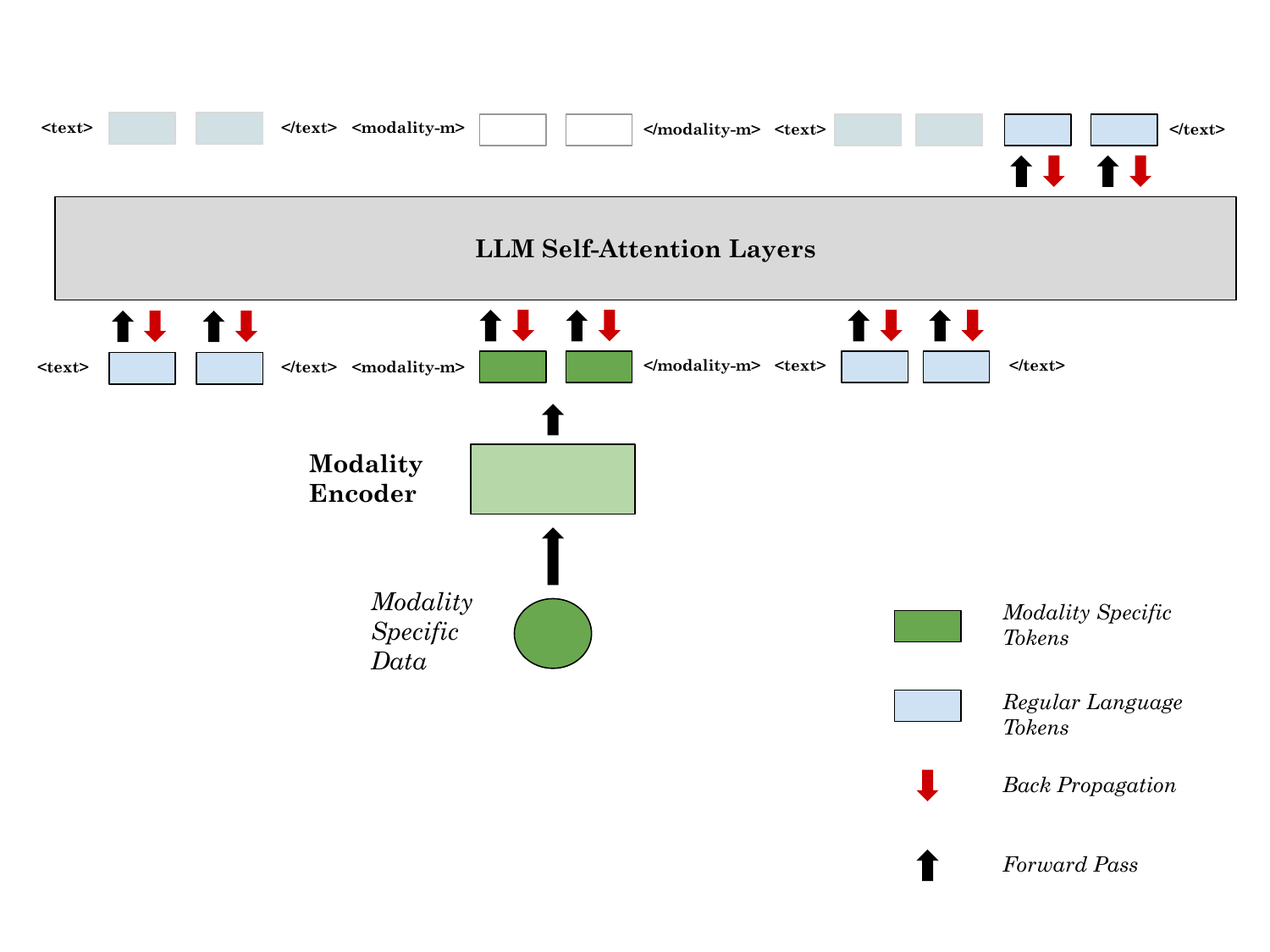}
\caption{Kosmos-I \cite{huang2023language} architecture where input from any modality is tokenized with a joint embedder using their encoded embeddings and subsequences of tokens are decorated with special indicative tokens according to their modality.}
\end{figure}

\section{Entity Embeddings : Everything as a Modality}

Reflecting on these developments and the necessities for better artificial cognition, it is evident that multi-modality will be one of the dominant attributes of the next models. In case the current transformer centric tensorial paradigm continues, we can postulate that it will be somehow in the generic tokenized form as the previously referenced architectures. Pure natural language will probably be the binding medium of numerous modalities and the cognitive engine for these models as in the case of human intellect \cite{vygotsky2012thought}\cite{carruthers2002cognitive}\cite{carruthers2002modularity}\cite{langland2014inner}. Towards such an era, which we refer as “omni-modal” in this paper, we also hypothesize that everything including pieces of text which belong to a common semantic entity can also be imagined as a modality.

\textbf{Entity as Modality} : Named Entity Recognition (NER) is a well established application of natural language processing \cite{nadeau2007survey}\cite{thomas2020deep}. State-of-the-art algorithms are capable of identifying most intricate and use case specific entities such as dosage of a medication in a medical document or a verdict in a legal text \cite{chirila2019named}\cite{leitner2019fine}. The human brain processes and stores multi-dimensional information by assigning them to latent meta-entities, somehow similar to NER categories, which provides high compression and neural efficiency\cite{rosch1978principles}\cite{patterson2007you}. Even with continuously increasing context lengths, current LLMs will always continue to struggle processing vast knowledge, which necessitated the development of Retrieval Augmented Generation (RAG) concept \cite{lewis2020retrieval}\cite{pan2022end}. Furthermore, even a quasi-infinite context length can be attained hypothetically, certain types of inefficiencies in self-attention transformer architectures emerge to process this information \cite{liu2023lost}.

One particular solution to adapt transformer based LLMs to process much larger amounts of information and allow modularity is to reformulate almost any entity including textual sequences as a modularity. For instance, textual references to a region in the document can be identified as a “location entity”, and rather than encoding individual tokens of the raw text, a pre-trained geographical encoder can generate modality tokens to feed in the joint linguistic space. This geography modular encoder may be separately trained including cartographic data like OpenStreetMap (OSM) including all geometrical and infrastructural information \cite{haklay2008openstreetmap}, a textual description from a source like Wikipedia and so on. In this modality specific formulation, even LLM does not possess any information about a new place would be able to extract and interpolate data from the embeddings, as it would be jointly trained generically to decode embeddings of geography modularity. As it can be seen, thinking of textual entities as a modality allows compressing huge amounts of information in several tokens, which might be one of the solutions to context limitation. In addition, the modality encoders and LLMs can be jointly trained to yield functional approximations hard to achieve by regular natural processing, such as estimating distance between different geographical locations as their embeddings include coordinates or assessing public transportation options as embeddings have processed their OSM data.

\section{Potential Examples of Entities as Modalities}

In this regard, we provide several illustrative examples that can be imagined in an omni-modality LLM era. Note that, in the presence of numerous modalities, the potential applications are vast, however only few exemplary proposals are made in order to shed light on the transformative promise. The regular natural modalities such as image, video, audio, signals shall also be included as interleaved tokens as in \cite{huang2023language} in this omni-modal context.

\textbf{Geospatial Entities} : The location is one of the most important aspects of human life and with the currently available colossal sum of cartographic data from layouts of power lines to highways, from amenities to natural features; LLMs have expansive potential to process these. However, these data are too voluminous to put in a single context. Even a RAG mechanism would yield instances of data too large to fit in a single context. Moreover, the inherent raw representations of these data in geographical coordinates make it difficult for LLMs to draw conclusions on proximity, path planning etc. The necessity of incorporating geographical features in LLMs is also discussed by various researchers \cite{janowicz2020geoai}\cite{rao2023building}.  

A scenario where geographical entities are processed in a modality specific manner is illustrated in Fig. 4. As it can be seen from the figure, to be encoded information by the geospatial modality encoder can include coordinate polygons of borders of the location, administrative regions, cartographic descriptions of amenities like museums, restaurants etc., public transport data, geometries of roads, streets, railways etc. and so on. One can see that if not explicitly provided by the user along with its raw data, the omni-modal LLM has to automatically assess the modality of the tokens in inference. We discuss this issue in the next section in more depth. Another aspect we would like to highlight is the possibility of using entity modalities in a nested fashion, both for the same modality as the parent entity and other modalities. As in Fig. 4, the data given to a geospatial encoder can contain other geospatial entities along with other modalities such as imagery. In the next section, we also address this recursive nature of omni-modality which allows for a much more comprehensive representation.

\begin{figure}[h]
\centering
\label{fig:fig_4}
\includegraphics[width=0.65\linewidth]{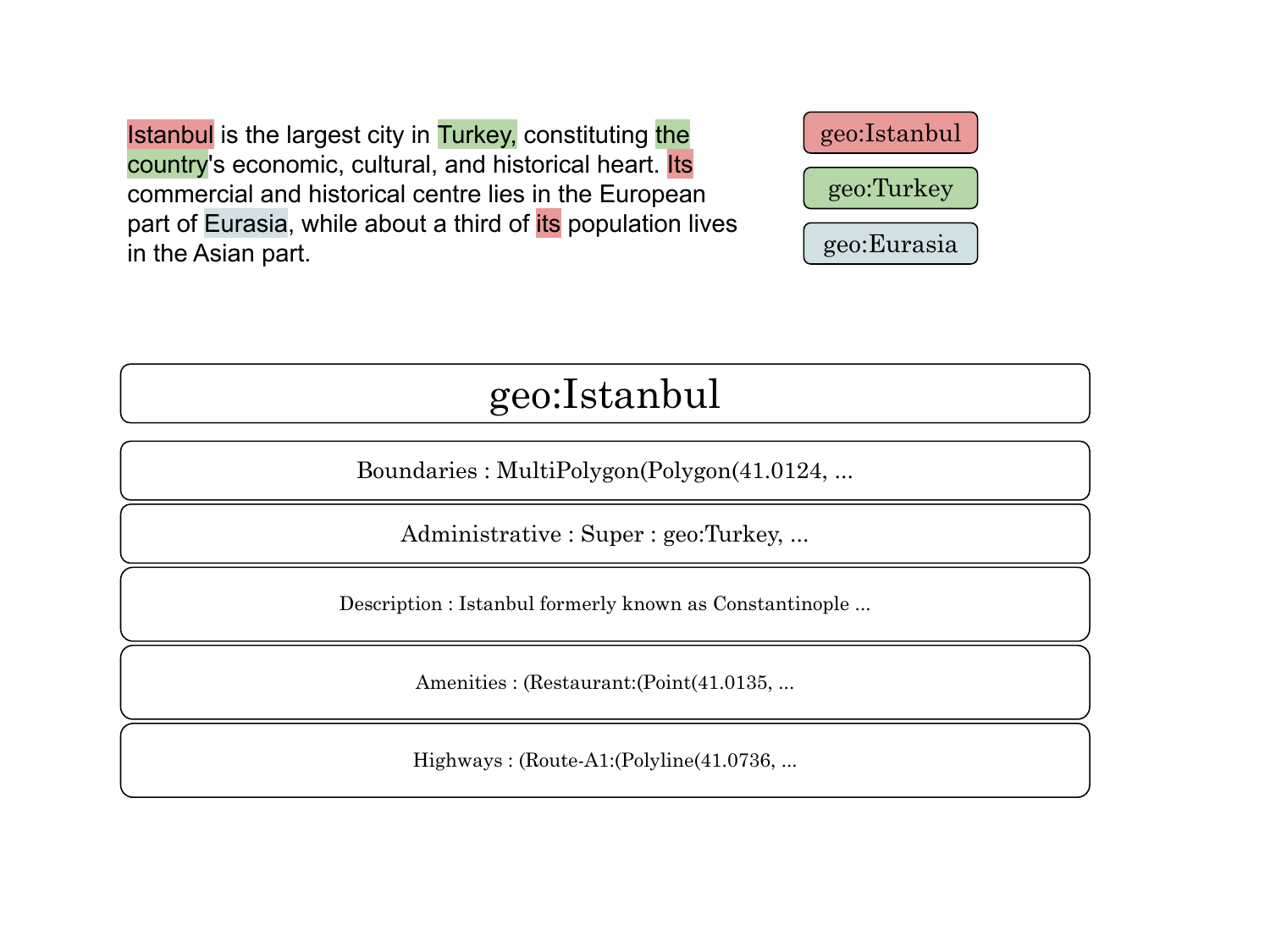}
\caption{An example of a textual passage where omni-modal LLM processes geographical entity tokens. The geography specific encoder is potentially fed with a multitude of cartographic and semantic data.}
\end{figure}

If geospatial encoder and binding omni-modal LLM are formulated and trained properly, the language model can gain capacity to interpolate distances and to a wider extent, notions of proximity, reachability by different modes of transport and natural topography. As current purely textual LLMs can not process algebraic or geometric tasks with sufficient performance, such alignment is necessary for geospatial context.  

\textbf{Corporations} : When we think of an object oriented, entity centric paradigm, one of the first categories that come to mind are persons or institutes like companies. For demonstrative purposes we give corporations as an example. At this point, it is important to mention the granularity of modularization. We talk about the issue of inclusiveness-feasibility trade-off for omni-modal networks in the next section. One can see that, when the aim is to produce a generic omni-modal LLM, the finer you divide your categories, the harder the management gets. This includes templates for raw data, data access, dataset curation and training regimes. For instance, for such a generic setting, one can choose to define “establishments” as an entity, which also includes commercial corporations, universities etc. Direct conclusions on this issue are out of scope of this work. 

As it can be seen in Fig. 5, a token for a specific corporation can be encoded with a specific modularity encoder which also can process tokens of other modalities including geolocation, people, dates etc. 

\begin{figure}[h]
\centering
\label{fig:fig_5}
\includegraphics[width=0.65\linewidth]{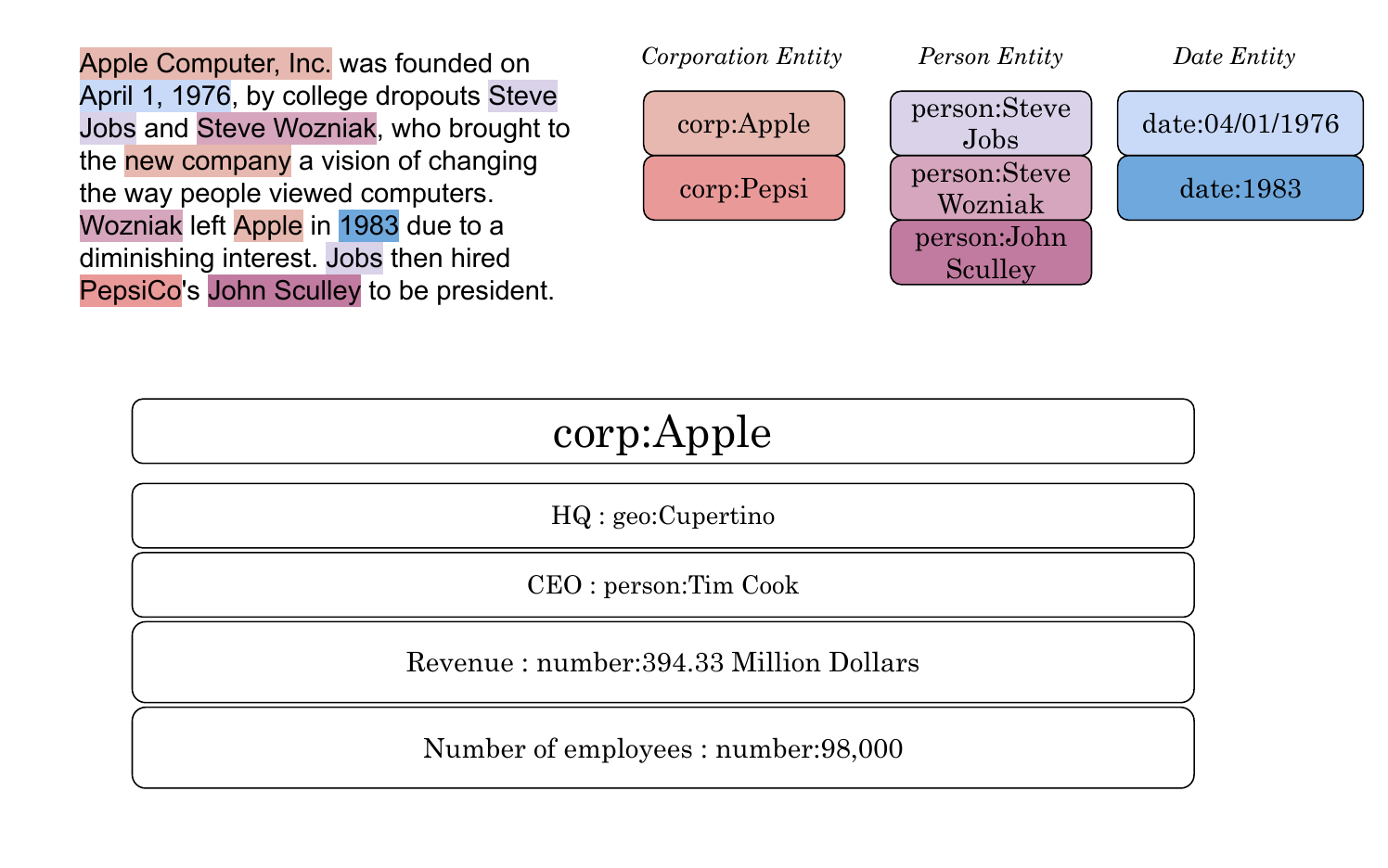}
\caption{Corporation as a modality. Note that, the raw data to be encoded by corporation specific modality encoder includes other modalities like numbers, people and geolocation.}
\end{figure}

\textbf{Numbers} : It is a well established fact that the current form of LLMs can not process numerical values efficiently to perform algebraic tasks and extrapolate \cite{imani2023mathprompter}\cite{yu2023metamath}. Approaching numbers as an entity might solve this problem under this omni-modal context. In case the number modality encoders and omni-modal LLM are trained accordingly, efficient algebraic manipulation, numeric extrapolation and cognition of quantitative relationships can be gained.

\textbf{Dates} : Concept of time is a field where a modality specific encoding can augment cognitive prowess. Date and time inherently have algebraic nature, so numerical limitation of current LLMs to construct relationships exists for them as well. One particular aspect of temporal entities is that they can be represented as in a numerical date-time format, a purely textual description or a mixture of both : “04/01/1976”, “70s”, “Middle Ages”, “14:50”, “before midnight” are all texts which are candidates to be encoded with temporal modality. As it can be seen, it imposes challenges both to define a proper boundary for the definition of time and to define a universal encoder to process this. However, it is evident that representing calendar and date-time concepts as modality would provide augmented temporal cognition.

\section{Towards Omni-Modal Era : Prospective Challenges and Rewards}

In addition to naturally distinct modalities such as image or audio, representing diverse references in texts as different modalities exhibits prospective capacity along with foreseeable challenges. First of all, most of the previously discussed architectures with interleavable tokens consider only encoding modalities, where generating modality tokens is not considered. This is logical as generation of high dimensional modalities such as image or video with sufficient quality is a tremendously complex aspect in a generic, universal setting. First and foremost, the next direction that the research will probably take will be the high-quality token generation incorporated into aforementioned architectures. In an entity-centric omni-modal perspective having modality-specific decoders shall be necessary.

For the entity-as-a-modality concept explained in this paper, the first challenge is the classification of entities in datasets, both during training and inference. Unlike modalities like images, the proposed approach includes modalities which are not explicit. As these are intangible modalities needed to be extracted from the text (or multi-modal data including any modality as image along with the text), a purposeful mechanism is required. Given the prowess of current LLMs in NER assessment this seems as a minor problem \cite{wang2023gptner}, however for such a universal modality processing concept things can become more challenging. Proper mechanisms are needed to modify datasets at the beginning, which might allow for self-supervison in later stages. 

Encoders of the conceptual entities such as geolocations or numbers should be carefully designed and trained such that desired abilities can be projected into / out from the LLM, like assessing distances from coordinates or performing calculus.

As mentioned previously, another challenge is the inclusiveness-feasibility trade-off on the definition of modalities. In the hierarchical semantic nature of things, it is hard to draw borders : finer concepts would require an exponentially larger number of modality encoders, dataset curation and training regimen, whereas separate categories more effectively. In contrast, coarser and more generic definitions of concept allow managing a lower number of specialized components and training dataset. 

One particular issue is about considering how to store, access and process the meta-data. Before embedding a specific entity, such as a company, all the related updated information such as revenue or number of employees should be delivered to the modality-specific encoder.

Perhaps one of the most powerful aspects of the proposed approach is the ability to encode modalities in a nested fashion. Description of a geolocation can include image, text, other geolocations and so on. This infinitely iterative and cross-modality architecture offers substantial cognitive power but requires overcoming related challenges in encoder design and training. 

Considering all, it seems such a formulation or a similar approach might be the only solution to context limitation, which also can not be solved properly with RAG. At the end, from a perspective this type of a modular architecture aims to distribute information to a set of composable encoders. Another advantage of this modularity is the allowance for new information injection. When the data of an entity changes such as fiscal data of a company the encoder universally embeds the modified data in the raw form. This also allows us to add many more new companies without any architectural change.  

\section{Conclusion}

The multi-modal LLMs with interleavable tokens have been introduced recently which projects all data into the joint linguistic embedding space. In this paper, we have hypothesized that in addition to explicitly distinct modalities like image or audio, sequences of texts which refer to semantic entities can also be formulated as modalities. These entities can be anything which belong to a common structured semantics : geographical location, time, people, numbers etc. With proper encoder architectures and training frameworks, they can project information into the common linguistic dimension of the binding LLM such that complex abilities on extrapolation and reasoning can be gained. This forecasted direction of the transformer based LLMs is referred to as the “omni-modal era”. Such an approach also allows the hierarchical and iterative encoding of the entities. Another advantage of this type of formulation is the information distribution on separate encoders and raw meta-data and modularity, which is promising to solve context length limitation, constraints of RAG and information update. In addition to prospective advantages of the omni-modal approach, challenges for such an era are also discussed.

\end{document}